\documentclass[10pt]{article}
\usepackage[a4paper, total={6.5in, 10in}]{geometry}
\usepackage{wrapfig}

\usepackage{times}
\usepackage{epsfig}
\usepackage{graphicx}
\usepackage{amsmath}
\usepackage{amssymb}

\usepackage[breaklinks=true,bookmarks=false]{hyperref} 

\date{}
\begin{document}

\title{YOLO-Z: Improving small object detection in YOLOv5 for autonomous vehicles}


\author{Aduen Benjumea\thanks{Visual Artificial Intelligence Laboratory, Oxford Brookes University, UK} 
\and Izzeddin Teeti\thanks{Autonomous Driving \& Intelligent Transport Group, Oxford Brookes University, UK} 
\and Fabio Cuzzolin\footnotemark[2] 
\and Andrew Bradley\footnotemark[1]
\and \tt\small17065125@brookes.ac.uk, 19136994@brookes.ac.uk,
\and \tt \small fabio.cuzzolin@brookes.ac.uk, abradley@brookes.ac.uk
}
  


\maketitle

\begin{abstract}
As autonomous vehicles and autonomous racing rise in popularity, so does the need for faster and more accurate detectors. While our naked eyes are able to extract contextual information almost instantly, even from far away, image resolution and computational resources limitations make detecting smaller objects (that is, objects that occupy a small pixel area in the input image) a genuinely challenging task for machines and a wide-open research field. This study explores how the popular YOLOv5 object detector can be modified to improve its performance in detecting smaller objects, with a particular application in autonomous racing. To achieve this, we investigate how replacing certain structural elements of the model (as well as their connections and other parameters) can affect performance and inference time. In doing so, we propose a series of models at different scales, which we name `YOLO-Z', and which display an improvement of up to 6.9\% in mAP when detecting smaller objects at 50\% IOU, at the cost of just a 3ms increase in inference time compared to the original YOLOv5. Our objective is to inform future research on the potential of adjusting a popular detector such as YOLOv5 to address specific tasks and provide insights on how specific changes can impact small object detection. Such findings, applied to the broader context of autonomous vehicles, could increase the amount of contextual information available to such systems.   
\end{abstract}

\section{Introduction}

Detecting small objects in images can be challenging, mainly due to limited resolution and context information available to a model \cite{cao2018feature}. Many modern systems that implement object detection do so at real-time speeds, setting specific requirements in computational resources, especially if the processing is to happen on the same device that captures the images. This is the case for many autonomous vehicle systems \cite{Culley2020}, where the vehicle itself captures and processes images in real-time, often to inform its next actions. In this context, detecting smaller objects means detecting objects farther away from the car, thus allowing earlier detection of such objects, effectively expanding the detection range of the vehicle. Improvements in this specific area would better inform the system, allowing it to make more robust and viable decisions.

Due to the nature of object detectors, the details of smaller objects lose significance as they are processed by each layer of their convolutional backbone. In this study, by `small objects', we refer to objects which occupy a small pixel area in the input image.

Efforts have been made to improve the detection of smaller objects \cite{Nguyen2020}, but many revolve around directing the processing around a specific area of the image \cite{Singh, Singh2018, Singh2018a} or are focused around two-stage detectors, which are known for achieving better performance at the cost of inference time, making them less suited for real-time applications. This is also the reason why so many single-stage detectors have been developed for this type of applications \cite{Wu}. Increasing the input image resolution is another obvious way to bypass this issue which results, however, in a significant increase in processing time.

YOLOv5 is a very popular single-stage object detector \cite{glenn_jocher_2021_4679653} known for its performance and speed with a clear and flexible structure that can be broken down, adjusted and built on a very widely accessible platform. Many of the systems that apply this architecture and attempt to optimise it, however, they mainly rely on adjusting specific parameters or augmenting their training set to improve performance \cite{Yang2020}, without much consideration for structural changes to the model itself to better adapt it for a specific use case. While YOLOv5 is a potent tool, it is designed to be a general-purpose object detector and therefore is not optimised to detect smaller objects.

This study proposes ways in which YOLOv5 can be modified to better perform on a given system in terms of small object detection, with clear real-world implications \cite{Culley2020}. We consider, in particular, the case of an autonomous racing vehicle that needs to detect differently coloured cones to drive around a track. We will discuss the effects of different techniques and propose modified models able to perform this task better while maintaining real-time processing speeds. The contributions of this paper are:

\begin{enumerate}
    \item A modified model of YOLOv5 specifically designed for better detections of small objects.
    \item Proposing a methodology to modify the structure of YOLOv5 to improve performance in a particular task. This is done in an experimental manner, analysing the different elements that make YOLOv5.
\end{enumerate}


\section{Related work}

This study aims at refining the already existing YOLOv5 model to deal with the many problems associated with small object detection. This task is a complex area of machine learning that very quickly escalates in complexity as requirements evolve. To work with such systems, it is essential to understand the bases upon which they are built, the many different technologies and techniques that form the current state of the art and the related use cases.

\subsection{One-stage vs two-stage objects detectors}

We know we can classify object detectors into two categories: one-stage and two-stage detectors \cite{Nguyen2020}. The latter typically decomposes the detection task into (i) region proposal generation and (ii) classification, as is the case with Faster R-CNN and its predecessors \cite{Girshicka, girshick2015fast, ren2016faster}. While there have been efforts to improve the small object detection ability of such models \cite{Chen2016}, a lot of the attention has been put on performance regardless of inference time. Two stage detectors have however improved significantly over time by streamlining their structure and data flow. 

\subsection{The YOLO family}

As a family of object detectors, YOLO takes this idea a step further and has grown very popular over the last few years. With YOLOv1 \cite{Redmon2015}, object detection is presented as a regression task, thus simplifying the networks and allowing us to build faster models that can be used in real-time. Later versions of YOLO improve different aspects of the model \cite{Redmon2015, Redmon2016, Redmon2018, Bochkovskiy2020}. Most notably, much effort has been spent on the backbone through the different versions. This begs the question: What potential is there still untapped if changes to an isolated element can have such an impact?

YOLOv5 \cite{glenn_jocher_2021_4679653} was released very shortly after YOLOv4 \cite{Bochkovskiy2020}. Despite its name, the authors are not directly related, and there have been discussions on whether it is fair to call YOLOv5 a successor of YOLOv4. This implementation provides similar performance to YOLOv4 and shares the same design. The main point of attention is the fact that it is fully written in the PyTorch framework \cite{PyTorch2019} as opposed to using any form of the Darknet framework \cite{darknet13} and has a focus on accessibility and use in a wider range of development environments. Additionally, the models in YOLOv5 prove to be significantly smaller, faster to train and more accessible to be used in a real-world application.

\subsection{Systems using and modifying YOLOv5}

YOLO has been used in many applications requiring the detection of objects. In safety helmet detection systems \cite{Zhou2021}, for instance, YOLO can be adjusted and implemented in series with the rest of a system. Similarly, face masks detectors have been seen at the entrances of metro stations \cite{Yang2020}. Both of these applications do a good job at exploiting the benefits of YOLO for the detection of smaller objects \cite{Pham2017}, but do not go as far as modifying the architecture.

Other systems that do make an effort to optimize YOLOv5 do so in a limited fashion. Once again, mask detectors \cite{Liu} have been proposed that leverage anchors generated and data augmentation to fit a model to the use case better. More complex systems for helmet detection \cite{Jia2021} also do a great job at leveraging the contextual information around small objects to isolate them and facilitate their detection. However, their approach is not quite universally applicable and comes at the cost of introducing a two-step process.

Typical adjustments to the internal structures of the model are surface-level. In a recent apple detection system \cite{Yan2021}, the backbone of YOLOv5 is slightly modified to simplify it, which offers the potential to adapt to the system's requirements and one that opens the way for additional changes. If a single backbone element is modified, more drastic changes can be applied for additional effects.

\subsection{Small object detection}

Some effort has been put into developing systems which direct the processing towards certain areas of the input image \cite{Singh, Singh2018, Singh2018a}, which allows us to adjust resolution and therefore bypass the limitation of having fewer pixels defining an object. This approach, however, is better suited for systems that are not time-sensitive, as they require multiple passes through a network at different scales. This idea of paying more attention to specific scales can nevertheless inspire the way we treat certain feature maps.

Additionally, a lot can be learned by looking at how feature maps can be treated instead of just modifying the backbone. Different types of feature pyramid networks (FPN) \cite{Lin2016, Tan2019, liu2018path} can aggregate feature maps differently to enhance a backbone in different ways. Such techniques prove to be rather effective. 

\subsection{Autonomous vehicles}

Within autonomous driving, object detection can provide valuable contextual information about the vehicle's surroundings and heavily inform its decision making process \cite{RobotCarDatasetIJRR, Culley2020}. In this case, smaller objects translate to objects further away, meaning a more complete context for the system to make use of. These  systems heavily focus on inference time, sacrificing performance if needed, but work can be done to improve them at minimal cost. Performance in this field is critical, as a small improvement in this system can greatly impact the entire vehicle. A common requirement in this area is for detectors to be single-stage \cite{Wu}, for the simple reason that fewer steps and transitions between them often translates into fewer resources needed.

\section{Methodology}

YOLOv5 provides four different scales for their model, $S$, $M$, $L$ and $X$ which stand for Small, Medium, Large, and Xlarge, respectively. Each of these scales applies a different multiplier to the depth and width of the model, meaning the overall structure of the model remains constant, but the size and complexity of each model are scaled. In Our experiments, we apply changes to the structure of the models individually across all the scales and treat each one as a different model for the purposes of evaluating their effect.

To set a baseline, we trained and tested the unmodified versions of the four scales of YOLOv5. We then tested changes to these networks individually in order to observe their impact separately against our baseline results. The techniques and structures that did not appear to contribute to better accuracy or inference time were filtered out when moving to the next phase. We then attempted combinations of the selected techniques. This process was repeated, observing whether certain techniques complemented or diminished each other and adding more complex combinations progressively.

We first discuss the appropriate evaluation metric for our work (Section \ref{sec:metric}), and the dataset used for our investigation (Section \ref{sec:dataset}). We then move on to describe our plans to apply a number of model changes to be run under controlled circumstances (Section \ref{sec:changes}), logging and adjusting as we move through different stages.

\subsection{Evaluation metric} \label{sec:metric}

The original implementation of YOLOv5 provides compatibility with Microsoft Common Objects in Context (COCO) API’s  \cite{Lin2014} metrics at three different object scales (bounding box areas) and Intersection over Unions (IOU
), which proves useful for the purpose of this study. 
The way values at specific scales are calculated can give us a good indication of the performance of the model, but may be slightly inaccurate in extreme cases, which will not be a problem for the most part, but must be kept in mind. 

Since these metrics are only compatible with the COCO dataset by default, we have re-implemented this method in our testing code in order to obtain more valuable figures for our study while using any dataset. Our metric module will calculate values for \emph{large}, \emph{medium} and \emph{small} objects, in addition to the overall performance. The categorisation of objects into these three categories depends on the following thresholds: `small', if the object occupy an area less than 32 squared pixels, `large', if the area is more than 96 squared pixels, and `medium', for anything between the two thresholds. In other words, small $< 32^{2} <$ medium $< 96^{2} <$ large.

\subsection{Dataset and Experimental setup} \label{sec:dataset}

\begin{wrapfigure}{r}{0.45\textwidth}
    \centering
    \includegraphics[width=0.5\textwidth]{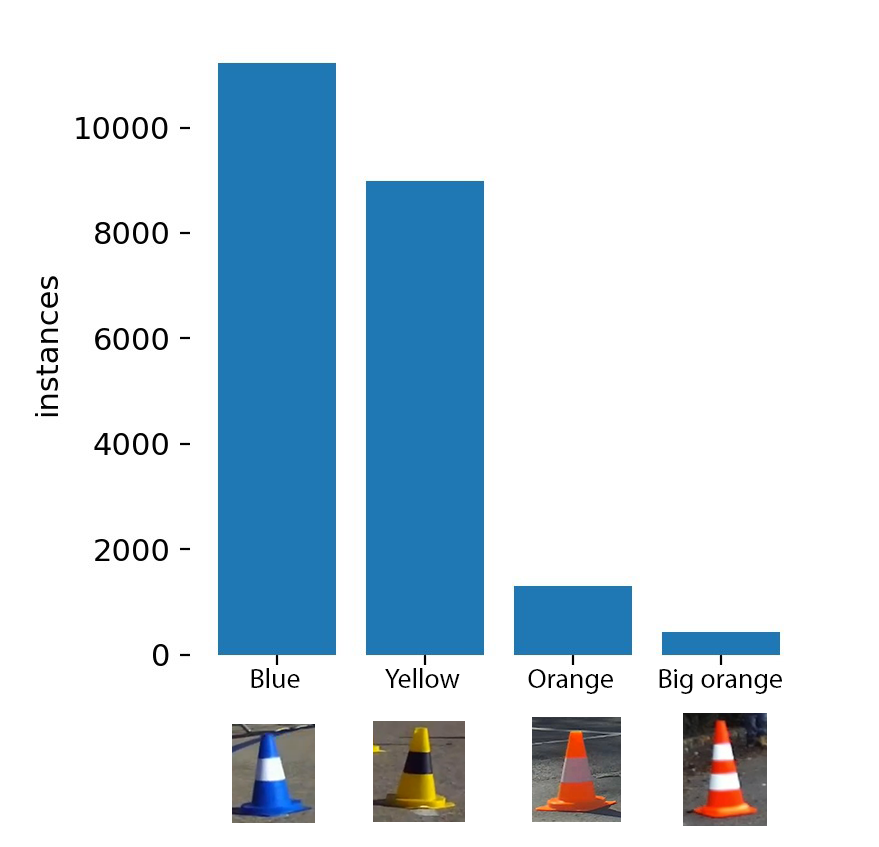}
    \caption{\footnotesize Dataset class instances}
    \label{fig:instance_count}
\end{wrapfigure}

\begin{wrapfigure}[9]{l}{0.45\textwidth}
    \vspace{-0.5cm}
    \centering
    \includegraphics[width=0.45\textwidth]{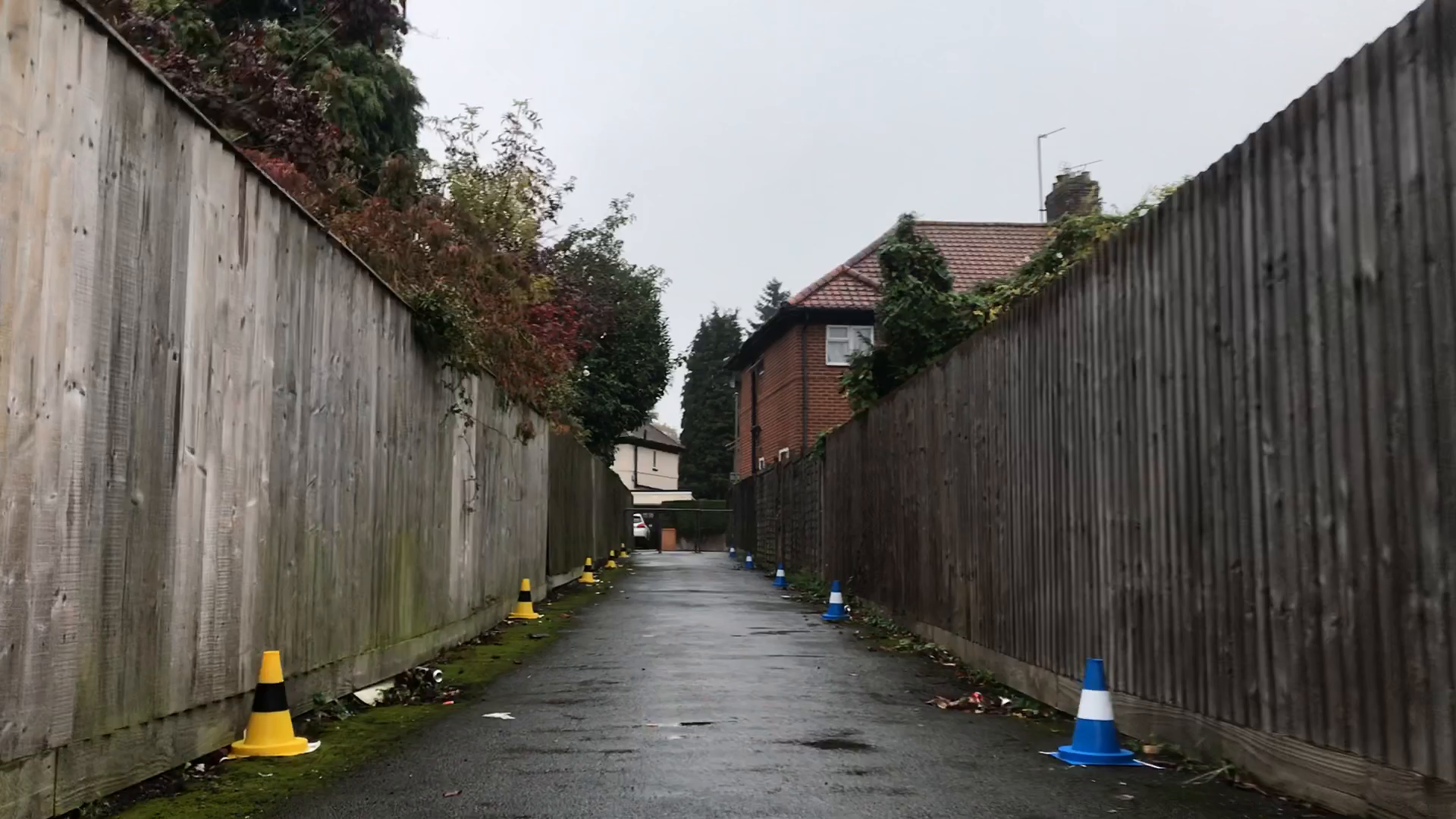}
    \caption{\footnotesize Sample image from dataset}
    \label{fig:sample_img}
\end{wrapfigure}

To train our models and inform our experiments we adopted a dataset of annotated cones from the perspective of an autonomous racing car. Its original purpose is to help plan a path for an autonomous racing vehicle based on the colours of the cones, knowing that there are a total of 4 classes (yellow, blue, orange and big orange cones) and close to 4,000 images (see Figure \ref{fig:instance_count}, \ref{fig:sample_img}). This dataset includes digitally augmented images \cite{Musat2021multi} and cases with challenging weather conditions. A dataset such as this one models more complex tasks in autonomous vehicles. Cones are themselves objects that we would find on the road and share many qualities with other objects such as of traffic signs in terms of size and position.

Although the dataset would benefit from a larger size, it is characterised by a very high object density, with over 30,000 labelled objects. Furthermore, looking at Figure \ref{fig:instance_count}, we can observe a very high bias towards the blue and yellow classes. This makes sense as they serve to mark the two sides of the racing track, but it does constitute an imbalance that will affect the overall results (see Section \ref{sec:experiments}). The performance on these classes will be taken into account when evaluating the models, namely by averaging the scores of the most prominent classes.

\begin{wrapfigure}[33]{r}{0.5\textwidth}
    \vspace*{-1.5cm}
    \centering
    \includegraphics[width=0.5\textwidth]{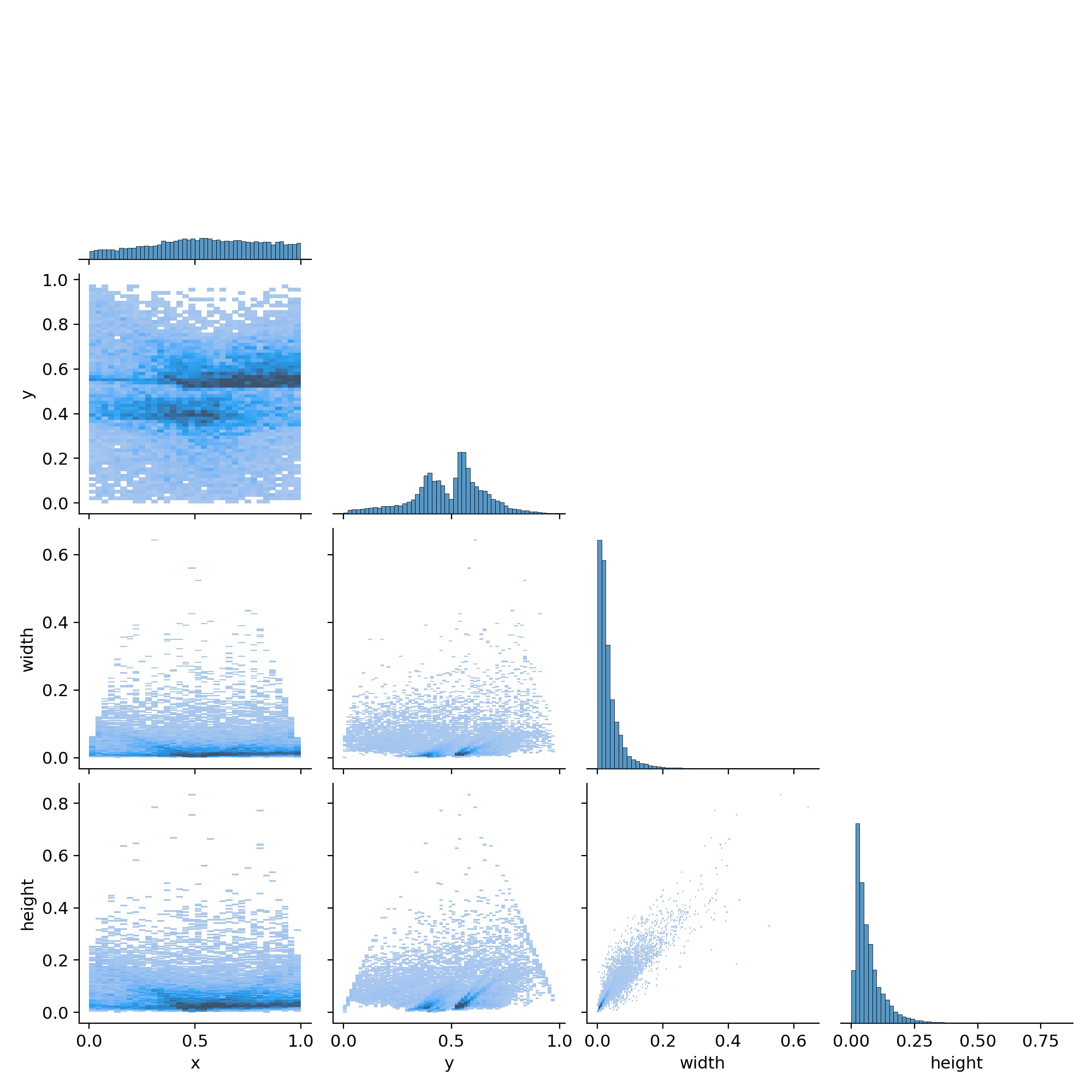}
    \caption{\footnotesize Relation between the position (in x and y value of the center point), width and height of instances of the dataset}
    \label{fig:correlogram}
    \hspace{1cm}
    \includegraphics[width=0.45\textwidth]{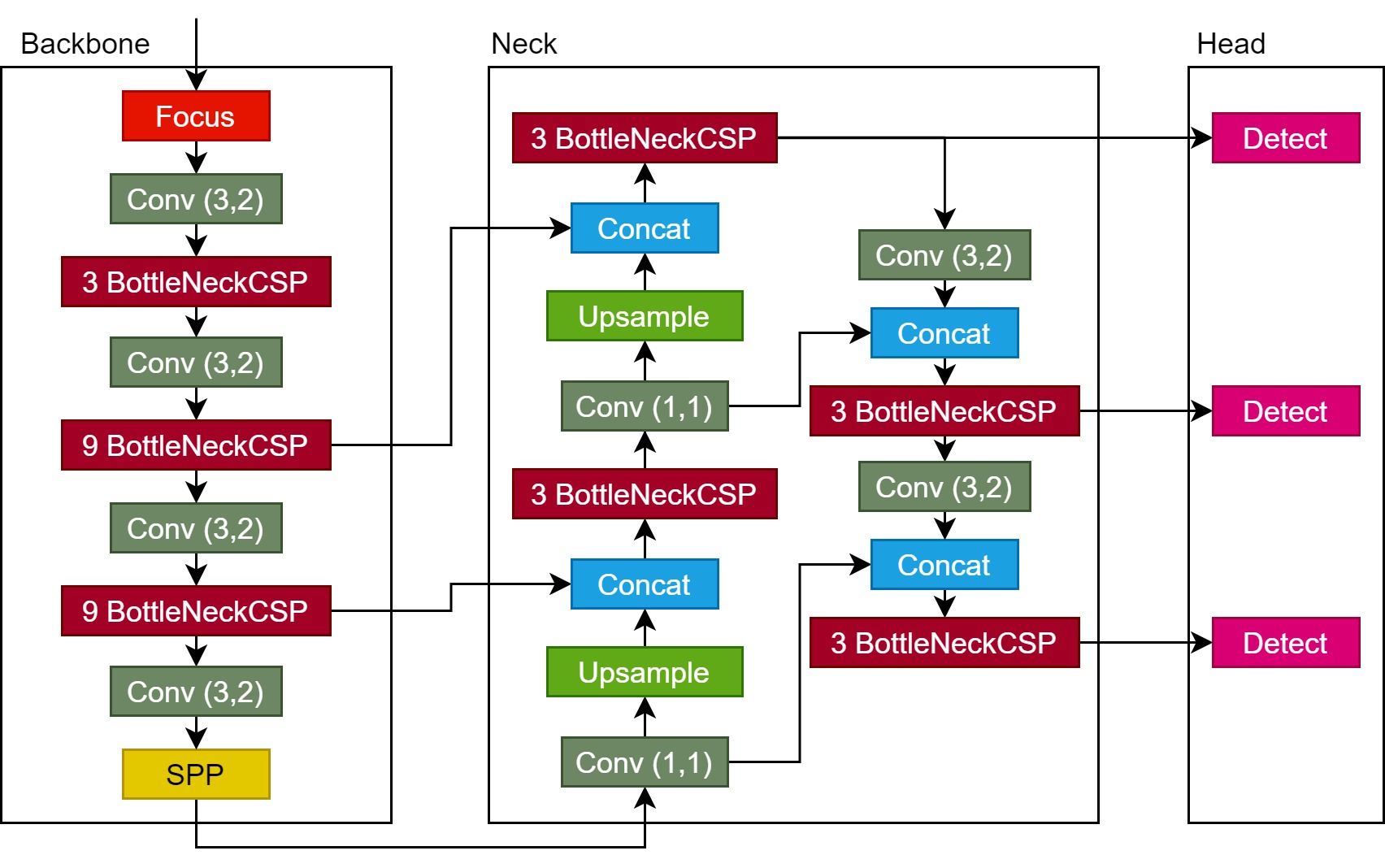}
    \caption{YOLOv5 default structure. In the text we refer to the elements of this architecture modified in our work.}
    \label{fig:yolo_diagram}
\end{wrapfigure}

Cones are naturally small objects already in comparison to other objects commonly found in the autonomous driving scenario, such as other vehicles or pedestrians. The correlogram (a chart of correlation statistics) in Figure \ref{fig:correlogram} shows the position, width, and height of the bounding boxes of the objects (cones) in the dataset. Our dataset features a high concentration of smaller object boxes, slightly elongated as to be expected because of perspective projection. This high proportion of small objects makes it beneficial for this type of study, as it largely overcomes the issue with a lack of such objects in other popular datasets including MS COCO \cite{kisantal2019augmentation}.

The dataset was split into training, validation and testing with a ratio of 65:15:20. The validation set then informs the training of the model, but is not as relevant as the other two, hence the lower size.

The training for all the experiments was executed in an environment that has 4 Nvidia GTX 1080 GPUs, each has 12 GB VRAM. For testing, however, we used a single GTX1080TI GPU with a batch size of 1, and an i7-6900K CPU working at 3.20GHz.

\newpage

\label{sec:changes}
\subsection{Proposed architectural changes} 

YOLOv5 uses a yaml file to instruct a parser how to build a model. We use this setup to write our own high-level instructions on how different building blocks of the model are built and with what parameters, hence modifying its structure. to implement new structures we arrange and give parameters to each building block or layer and instruct the parser on how to build it if necessary. In our words, we make use of the base and experimental network blocks provided with YOLOv5, while implementing additional blocks where needed to simulate the required structures.

\subsubsection{Backbone}

The backbone of a model is the element dedicated to taking the input image and extracting feature maps from it. This is a crucial step in any object detector, as it is the main structure responsible for extracting contextual information from the input image as well as for abstracting that information into patterns. 
We experimented with replacing the existing backbone in YOLOv5 with two separate options. ResNet \cite{He2015} is a popular structure that introduces residual connections to lessen the effect of the diminishing return we observe in deeper neural networks. DenseNet \cite{Huang} uses similar connections to preserve as much information as possible as it moves through the network. Implementing these structures requires breaking them down to their fundamental blocks and ensuring the layers communicate appropriately. This includes ensuring the right feature map dimensions, which at times requires slightly modifying the scaling factor for the width and depth of the model.

In both cases, it was important to avoid drastically deviating the number of layers from the original as to maintain a comparable complexity. Hence, ResNet50 was used and we downscaled DenseNet proportionally so it reattains its core functionality.



Additionally, YOLOv5 makes use of a Spatial Pyramid Pooling (SPP) \cite{He2014} layer in between the backbone and the neck. In our work, however, we have maintained this layer untouched.

\subsubsection{Neck}

We term `neck' the structure placed between the head and backbone (see Figure \ref{fig:yolo_diagram}) whose objective is to aggregate as much information extracted by the backbone as possible before it is fed to the head. This structure plays a major role in transferring small-object information by preventing it from being lost to higher levels of abstraction. It does this by upsampling the resolution of the feature maps once again so different layers from the backbone can be aggregated and regain influence on the detection step. 
\cite{liu2018path}.

In this work, we simplified the current Pan-Net \cite{liu2018path} to that of an FPN \cite{Lin2016} and replaced it a with biFPN \cite{Tan2019}. In both cases the neck retains a similar functionality, but varies in complexity and therefore the number of layers and connections required for their implementation.

\newpage
\subsubsection{Other modifications} \label{subsec:other}

\begin{wrapfigure}[32]{r}{0.55\textwidth}
    \vspace{1cm}
    \centering
    \includegraphics[width=0.5\textwidth]{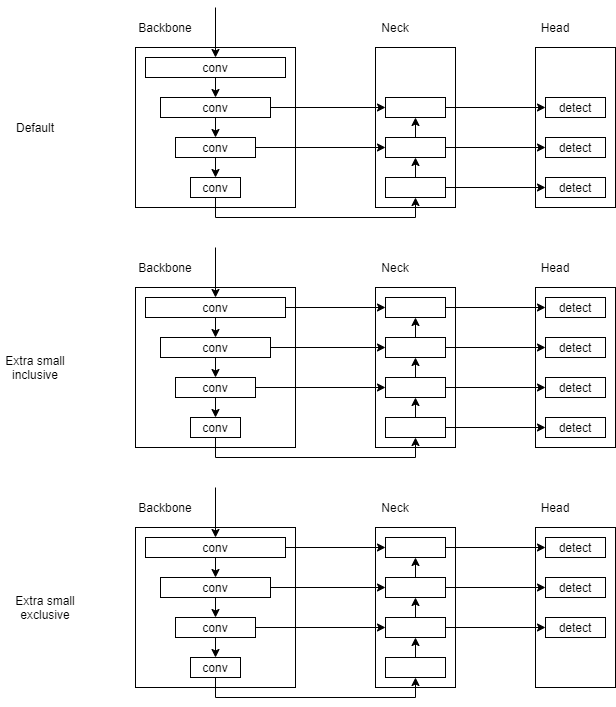}
    \caption{\footnotesize Example of how favoring smaller feature maps can be implemented in terms of structure both inclusively and exclusively.}
    \label{fig:XS_det_example}
\end{wrapfigure}

The head of the model is responsible for taking feature maps and inferring the bounding boxes and classes by taking in several aggregated feature maps from the neck. This structure can remain untouched, other than the parameters it receives, as it is a fundamental part of the model that does not have as much impact in small object detection as the aforementioned elements.
\\
There are, however, other elements that can have an impact on small object detection performance. Other than input image size, the depth and width of the model can be modified in order to change what aspect of the network the bulk of the processing goes towards. The way layers are connected can also be manually altered in the neck and head in order to focus on detecting certain feature maps.

In this study we explored the effect of redirecting the connections involving higher-resolution feature maps, in order for them to be fed directly to the neck and head. This can be done in an `inclusive' manner by expanding the neck to fit an extra feature map, or in an `exclusive' fashion by replacing the lowest-resolution feature map in order to fit the new one, Figure \ref{fig:XS_det_example} shows both options, as well as the default (original) layout. Making use of a higher resolution feature map would usually improve performance on smaller object at the cost of inference time and potentially detection of larger objects, similar to the effect of increasing input image size. We integrate this behaviour in the neck in these two ways to minimize the downsides while making the most out of its benefits.

Note that a number of parameters will have to be adjusted to the new structure, as the learning capabilities of the network can be affected. Mainly, the sizes of the anchor boxes applied in the head, which need to adjust to the resolution of the feature maps being used.


\begin{figure}
    \centering
    \includegraphics[width=0.4\textwidth]{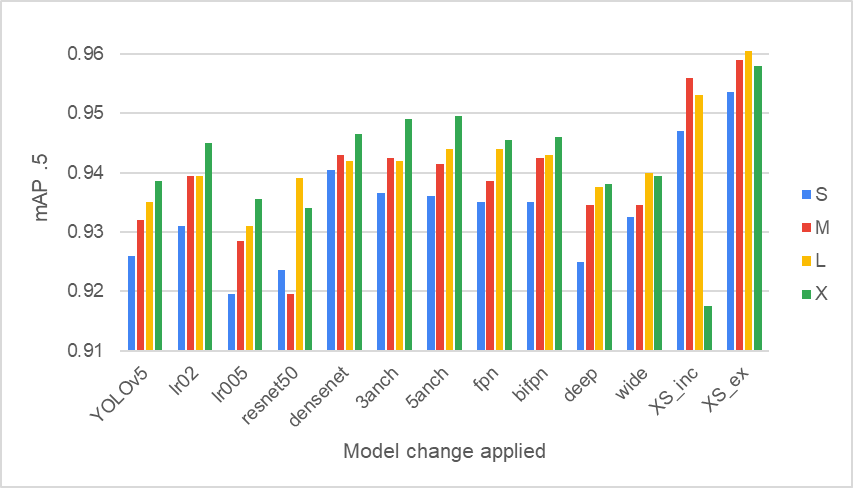}
    \includegraphics[width=0.4\textwidth]{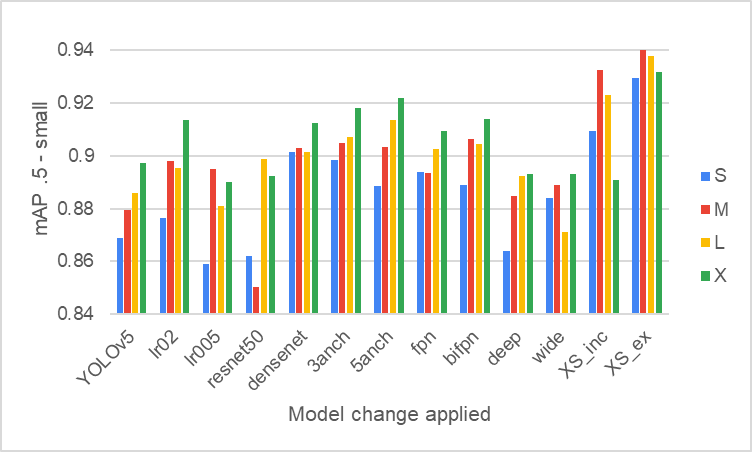}
    \includegraphics[width=0.4\textwidth]{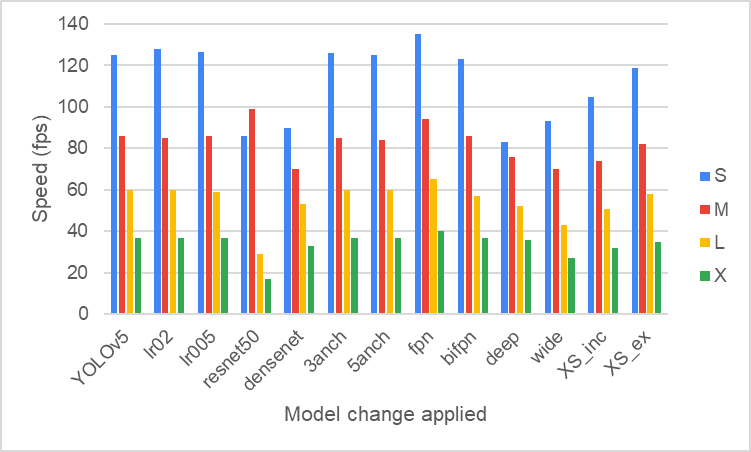}
    \caption{\footnotesize Results of applying individual architectural changes to YOLOv5 at each scale. We report the mAp at 50\% IOU across all objects sizes (top), the mAP at 50\% IOU for small objects only (middle), and the inference speed in frames per second (fps, bottom). \textbf{YOLOv5}: the baseline. \textbf{lr02}: changing the learning rate to 0.02. \textbf{lr005}: changing the learning rate to 0.005. \textbf{resnet50}: changing the backbone to that of ResNet50. \textbf{densenet}: changing the backbone to that of DenseNet. \textbf{3anch}: auto-generating 3 anchors per scale. \textbf{5anch}: auto-generating 5 anchors per scale. \textbf{fpn}: changing neck to that of FPN. \textbf{bifpn}: changing neck to that of BiFPN. \textbf{deep}: increasing the depth modifier to that of the next scale up (or equivalent). \textbf{wide}: increasing the width modifier to that of the next scale up (or equivalent). \textbf{$XS_{inc}$}: setting an extra small feature map inclusively (see \ref{subsec:other}). \textbf{$XS_{ex}$}: setting an extra small feature map exclusively (see \ref{subsec:other}).} 
    \label{fig:result_graphs}
\end{figure}

\newpage
\section{Results} \label{sec:experiments}

\begin{table*}[htbp]
  \centering
    \begin{tabular}{|l|l|}
    \hline
    Model & Features \\
    \hline
    YOLO-Z S & DenseNet, FPN, 3 anchors, extra small exclusive feature map \\
    YOLO-Z M & DenseNet, FPN, 5 anchors, extra small exclusive feature map \\
    YOLO-Z L & DenseNet, FPN, 5 anchors, extra small exclusive feature map \\
    YOLO-Z X & DenseNet, bi-FPN, 5 anchors, extra small exclusive feature map \\
    \hline
    \end{tabular}
    
    \caption{\footnotesize Modifications applied to YOLOv5 to achieve models of the YOLO-Z family. Each scale uses its YOLOv5 equivalent as a base. \label{tab:z-features}}
\end{table*}%

\begin{table*}[htbp]
\hspace*{-1cm}
  \centering
    \begin{tabular}{|c|rrr|rrr|rrr|}
    \hline
          & \multicolumn{3}{c|}{mAP .5} & \multicolumn{3}{c|}{mAP .5 small} & \multicolumn{3}{c|}{inference (ms)} \\
          \hline
    Scales & \multicolumn{1}{l}{YOLOv5} &
    \multicolumn{1}{l}{YOLO-Z} & \multicolumn{1}{l|}{difference } & \multicolumn{1}{l}{YOLOv5} & \multicolumn{1}{l}{YOLO-Z} & \multicolumn{1}{l|}{difference } & \multicolumn{1}{l}{YOLOv5} & \multicolumn{1}{l}{YOLO-Z} & \multicolumn{1}{l|}{difference} \\
    \hline
    S     & 0.926 & \textbf{0.955} & 3.13\% & 0.869 & \textbf{0.925} & 6.44\% & \textbf{8} & 8.9 & 0.9 
    \\
    M     & 0.932 & \textbf{0.9605} & 3.06\% & 0.8795 & \textbf{0.9425} & 7.16\% & \textbf{11.6} & 14.3  & 2.7
    \\
    L     & 0.935 & \textbf{0.964} & 3.10\% & 0.886 & \textbf{0.9545} & 7.73\% & \textbf{16.6}  & 19.6  & 3 
    \\
    X     & 0.9385 & \textbf{0.9605} & 2.34\% & 0.8975 & \textbf{0.9465} & 5.46\% & \textbf{26.9}  & 30.6  & 3.7 
    \\
    \hline
    \end{tabular}%
  \caption{\footnotesize Comparing performance and inference time of YOLOv5 and YOLO-Z (optimal values for each scale in bold). \label{tab:z-performance}}
\end{table*}%


Note that we only show performance on the yellow and blue classes, as they are the best represented in the dataset according to Figure \ref{fig:instance_count}. (See supplementary material "Individual test results", Table 1).

\subsection{Influence of the backbone}

\begin{figure}
    \vspace{-0.5cm}
    \centering
    \includegraphics[width=0.4\textwidth]{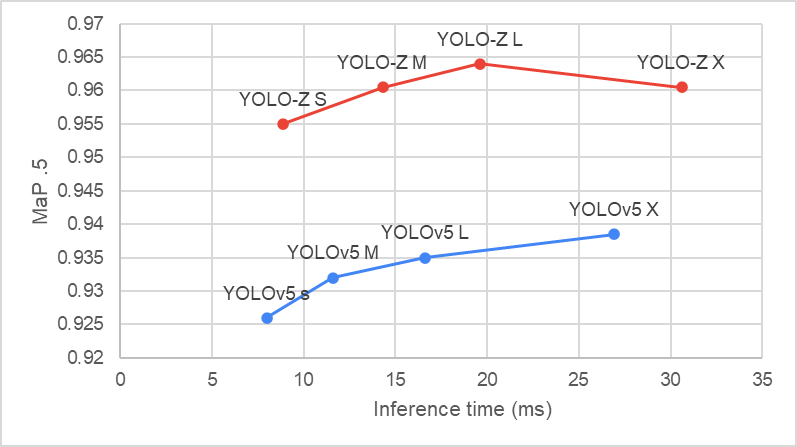}
    \includegraphics[width=0.4\textwidth]{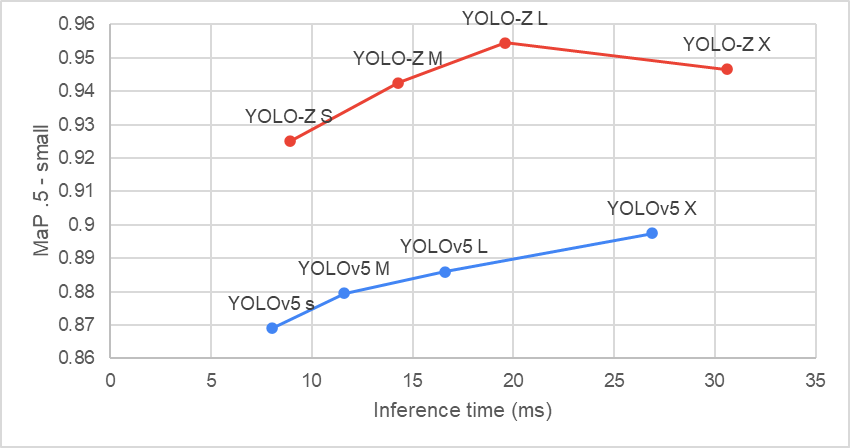}
    \includegraphics[width=0.4\textwidth]{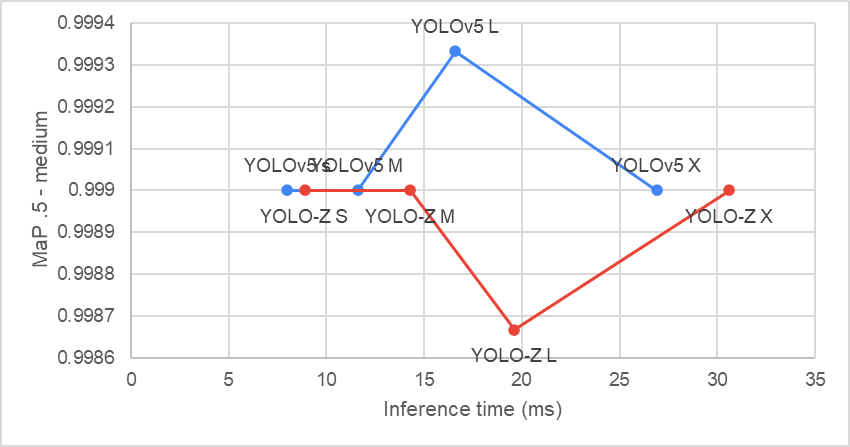}
    \caption{\footnotesize Performance comparison between the YOLOv5 and YOLO-Z families of models, plotting mAP (top), mAP for small objects (middle) and mAP for medium objects (bottom) against inference time (ms). Clearly the superior average performance of YOLO-Z is achieved at the smaller scale, while performance is stable and very close to 1 at the medium scale. \label{fig:fam_comparison}}
\end{figure}

A comparison of the performance of the two backbones (see Figure \ref{fig:result_graphs}) shows that DenseNet consistently exhibits a significant improvement at what appears to be a relatively low fixed increase in inference time (about 3 ms). ResNet not only seems to worsen performance in most cases, but its inference time is also significantly higher, leaving no reason to consider it further at this stage. Our conclusion is that DenseNet is therefore a better fit, in general, for small scale object detection. In the smaller scale models, 
this can be due to not having networks deep enough to reap the benefits of a ResNet backbone, while DenseNet does a good job at preserving feature maps' details.

\subsection{Influence of neck architecture} 

Using an FPN only outperforms bi-FPN at the $S$ scale (see Figure \ref{fig:result_graphs}).  In the latter case inference time remains fairly comparable to that of the original YOLOv5 neck, 
which is not too surprising given their similarities. This might suggest that simpler models benefit from keeping the feature maps relatively untouched, while other scales require extra steps to adapt to the added processing of the feature maps and eventually outperform the former.

\subsection{Feature maps} 

In our experiments, redirecting what feature maps are fed to the neck and head had the most significant impact among all techniques. Excluding the lowest resolution feature map in order to replace it with a higher-resolution one (\emph{XS\_ex} in Figure \ref{fig:result_graphs}) 
proved particularly effective. This can be attributed to the fact that, after including a higher resolution map in the head, small objects end up occupying more pixels and having therefore more of an influence, rather than being `lost' in the convolution stages of the backbone. Similarly, getting rid of the original lower-resolution feature map reduces the amount of processing needed and prevents the model from counteracting the level of detail provided by the higher-resolution map. This is likely a consequence of the dataset used having a very high density of small objects; performance will likely vary significantly in other applications. The only exception to this pattern seems to be that of the extra large scale ($X$), for which the improvement is not as significant and keeping the lower resolution feature maps actually appears detrimental to performance In comparison to the baseline. 

\subsection{Influence of the number of anchors}

Letting YOLO generate anchors based on the dataset provided proves effective in performance without affecting inference time. However, The magnitude of the effect and the number of anchors a model favors does seem to be affected by scale (\emph{anch3} and \emph{anch5} in Figure \ref{fig:result_graphs}). Again, the dataset used is relevant to this step as, in our tests, most cones will have a similar elongated shape on the y axis (see Figure \ref{fig:correlogram}). Other applications with objects varying more in shape might find different results. 

In terms of scale, the smaller models tend to benefit from fewer anchors, while the opposite is true for the larger scales. Namely, at the $S$ scale having 3 anchors outperforms having 5, while the gap reduces at the $M$ scale. 
Models $L$ and $X$, on the other hand, display instead a better performance at 5 anchors. This suggests that more complex or deeper models may indeed benefit from additional anchors or, in other words, may be more capable of taking advantage of the details additional anchors provide.

\subsection{Other factors}

\begin{wrapfigure}[28]{r}{0.4\textwidth}
    \centering
    \includegraphics[width=0.4\textwidth]{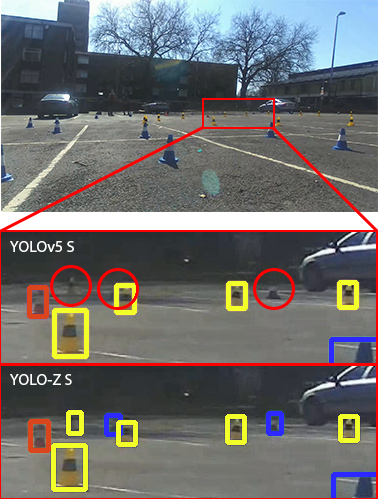}
    \caption{\footnotesize Visual demonstration of the improved detection results of YOLO-Z S (bottom) compared to YOLOv5 S (top) over a region of a sample image covering far away / small scale objects. Yellow and blue cone detections are shown as bounding boxes in the respective colours, detections missed by both models are shown as red boxes, detections missed by YOLOv5 but correctly identified by YOLO-Z S as red circles. One can observe that the improvement is most evident for farther away / smaller cones. \label{fig:vis_demo}}
\end{wrapfigure}

In addition to the other major structural changes, a larger learning rate proved to better leverage the models, but this can vary with the number of epochs the latter are trained for (\emph{lr02} and \emph{lr005} in Figure \ref{fig:result_graphs}). 

Interestingly, a wider model (higher width multiplier)  
showed to have a positive effect on the smaller scales as opposed to a deeper one (\emph{deep} and \emph{wide} in Figure \ref{fig:result_graphs}). The opposite is true for the $L$ scale. This might be due to the specific characteristics of our dataset, in which the objects to be detected form relatively simple patterns with rather similar features. Nevertheless, more testing would be needed to determine this, as this pattern is not continued in the extra large scale, where both modifications harm performance by the same amount. Additionally, these types of alterations do have a noticeable negative effect on inference speed, discouraging their use.

\subsection{Modified models}

Additional tests were carried out using various combinations of the aforementioned alteration techniques in order to seek models that further deviate from the originals but, at the same time, can further improve performance. 
We refer to this proposed family of models at different scales as YOLO-Z, short for `YOLO Zoomed' (see Table \ref{tab:z-features}).

A comparison of the performance of these new models shows that an FPN neck tends to outperform bi-FPN (compare YOLO-Z X with the other YOLO-Z models in Table \ref{tab:z-performance}) for scales where the opposite was previously true. Aside from this, we could only observe small variations across the models. An exception is the X scale, which seems to gain less from such changes and even with the use of a different neck structure does not deliver improvements as significant as with the other scales. 

This considered, YOLO-Z models achieved an average 2.7 performance increase in absolute mAP at 50\% IoU for all objects and an absolute improvement of 5.9 for small objects at the same IoU across all scales. This comes at the cost of an average 2.6ms increase in inference time.

\subsection{Discussion}

In our investigation of 
ways in which a popular object detector such as YOLOv5 can be adapted to better detect smaller objects, we were able identify architectural modifications delivering a clear improvement in performance compared to original at relatively little cost, as the new models retain real-time inference speed.

The context in which we have applied the proposed techniques, that of autonomous racing, is one that can greatly benefit from such an improvement. As we can see in Figure \ref{fig:vis_demo}, such changes do have a quantifiable impact on detection. In this work we have not only significantly improved the performance of the baseline model, but also identified a number of specific techniques that can be applied to any other application involving the detection of small or far away objects.

The net result is that models of the YOLO-Z family outperform those of the YOLOv5 class while retaining an inference time compatible with a real time application such as autonomous racing (see Table \ref{tab:z-performance} and Figure \ref{fig:fam_comparison}). This is especially true for the smaller objects which have been the focus of this study (Figure \ref{fig:fam_comparison}, middle), whereas the performance is stable for medium-sized objects (bottom).

Note that, while we have focused here on modifying the popular YOLOv5 model, the methods and techniques we explored 
have a potential to be developed into an entirely original model structure. 

Finally, while this study shows significant the empirical gains of the proposed architectural changes, the consistency and generality of the results could and should be further investigated.
For instance, the analysis
would greatly benefit from further testing with different datasets, and challenges that might for instance come from detecting traffic signs. While we have demonstrated the usefulness of the various techniques introduced, these can only be refined and better understood when applied to a varied set of circumstances and settings. Doing so would be a significant step towards a more robust solution to small object detection. Additionally, there many more directions and techniques that would fit nicely in this subject and which have not been considered in this study, but these will remain a subject of future study.


\section{Conclusions}

In this study, we have investigated the effects and rationale of different architectural and model alterations 
applied to the popular YOLOv5 object detector in order to improve its small-object detection abilities. We have 
validated these techniques in the
autonomous racing scenario, 
highlighting its specific needs and limitations, and outlining possible further research. Doing so has produced an original YOLO-Z family of models capable of delivering an improvement in the ability of detecting small objects (measured by the standard mAP at 50\% IoU) close to 6\%, while only increasing inference time by around 3 ms. Using these findings existing systems can be upgraded to better detect very small objects in situation in which current models cannot detect anything at all. This can extend an autonomous vehicle's detection range and perception robustness, leading to better planning and decision making strategies that can give an autonomous racing car an important edge.


\section{Acknowledgements}

Thanks go to the Autonomous Driving \& Intelligent Transport group at Oxford Brookes University, and the OBR Autonomous team for their input and assistance. The authors would also like to thank Salman Khan, Peter Ball, Tjeerd Olde Scheper, Matthias Rolf, Alex Rast, and Gordana Collier for their ongoing support. 

This project has received funding from the European Union’s Horizon 2020 research and innovation programme, under grant agreement No. 964505 (E-pi).

{\small
\bibliographystyle{ieee_fullname}
\bibliography{egbib}
}

\end{document}